\documentclass[a4paper,conference]{IEEEtran}

\usepackage{times}
\usepackage{epsfig}
\usepackage{graphicx}
\usepackage{amsmath}
\usepackage{amssymb}
\usepackage{multirow}
\usepackage{color}
\usepackage{url}
\usepackage{subfigure}
\usepackage[colorlinks,linkcolor=blue]{hyperref}

\newcommand{\tabincell}[2]{\begin{tabular}{@{}#1@{}}#2\end{tabular}}

\usepackage{cite}

\ifCLASSINFOpdf
\else
\fi

\begin{document}
\title{An Efficient End-to-End 3D Voxel Reconstruction based on Neural Architecture Search}

\author{\IEEEauthorblockN{Yongdong Huang, Yuanzhan Li, Xulong Cao \\ Siyu Zhang, Shen Cai$^{\ast}$, Ting Lu}
\IEEEauthorblockA{Visual and Geometric Perception Lab\\
Donghua University\\
Shanghai 201620, China\\
$^{\ast}$Correspondence: hammer\_cai@163.com}
\and
\IEEEauthorblockN{Jie Wang$^{1, 2}$}
\IEEEauthorblockA{$^{1}$Visual and Geometric Perception Lab\\
Donghua University\\
Shanghai 201620, China \\
$^{2}$The University of Manchester}
\and
\IEEEauthorblockN{Yuqi Liu}
\IEEEauthorblockA{School of Electronics and \\ Information Engineering\\
Tongji University\\
Shanghai, 201804, China}
}


%

\maketitle

\begin{abstract}
Using neural networks to represent 3D objects has become popular. However, many previous works employ neural networks with fixed architecture and size to represent different 3D objects, which lead to excessive network parameters for simple objects and limited reconstruction accuracy for complex objects. For each 3D model, it is desirable to have an end-to-end neural network with as few parameters as possible to achieve high-fidelity reconstruction. In this paper, we propose an efficient voxel reconstruction method utilizing neural architecture search (NAS) and binary classification. Taking the number of layers, the number of nodes in each layer, and the activation function of each layer as the search space, a specific network architecture can be obtained based on reinforcement learning technology. Furthermore, to get rid of the traditional surface reconstruction algorithms (e.g., marching cube) used after network inference, we complete the end-to-end network by classifying binary voxels. Compared to other signed distance field (SDF) prediction or binary classification networks, our method achieves significantly higher reconstruction accuracy using fewer network parameters.
\end{abstract}


%
\IEEEpeerreviewmaketitle

\section{Introduction}
\label{sec:1}
With the increasing development of three-dimensional (3D) deep learning, the task of 3D object representation and reconstruction has become a research hotspot. 
3D objects can be represented explicitly or implicitly. 
Common explicit representations include point clouds, meshes, and voxels, among others. 
The point clouds representation stores the position of each point, and may also contain color and normal vectors, etc.
Mesh establishes the connectivity between points and forms facets for rendering models.
Voxel is a dense grid representation that requires a lot of storage. 
In addition, the octree representation can be used to dynamically adjust the spatial resolution based on the local details of the object. 
Along with the above explicit representations, neural networks have accomplished various 3D tasks in the areas of computer vision and computer graphics. 
For example, PointNet~\cite{pointnet} first uses MLP layers to obtain the high-dimensional feature of each point, which is then used for object classification and segmentation.
Volumetric 3D convolutional neural network (CNN) is introduced in VoxNet~\cite{voxnet}, and is still used for object classification.
ONet~\cite{ONET} extracts latent vectors of one category of objects from their single-view images, point clouds, or voxels, and accomplishes reconstruction by predicting the occupancy of voxels in a reconstructed model. 

Among the implicit representations, the signed distance field (SDF) is the most popular since it is a continuous and probability-like representation.
For a 3D point, the closer it is to the surface of the object, the smaller the absolute value of its SDF.
Based on SDF representation, neural networks have achieved good performance in many tasks, especially in object reconstruction.
For example, NI~\cite{NI} first proposes to use a multi-layer perception (MLP) network to overfit the SDF for each individual object.
Therefore, neural compression (through storing network parameters) and reconstruction (through network inference and subsequent surface reconstruction) are achieved.
NGLOD~\cite{nglod} adopts the idea of local fitting to significantly improve the reconstruction accuracy for an individual model.
The reconstruction error of NI or NGLOD designed for an individual object is obviously lower than the above mentioned ONet designed for one category of objects.
However, the compression task is ignored by NGLOD, since the method requires storing a large number of latent vectors of grid nodes, possibly even more than the number of vertices and faces of the model itself.
Different from the reconstruction task of 3D models, the recent famous NeRF works~\cite{nerf}~\cite{nerf++}\cite{qiuxieNerf} encode 3D scenes, including shapes, texture and illumination, from a set of calibrated images. 
Although they also use MLP networks to predict opacity that is similar to SDF, their goal is nearly irrelevant to the neural compression and reconstruction of known 3D models, which is the focus of this article.
\begin{figure*}[t]
	\centering
	\includegraphics[width=0.9\textwidth]{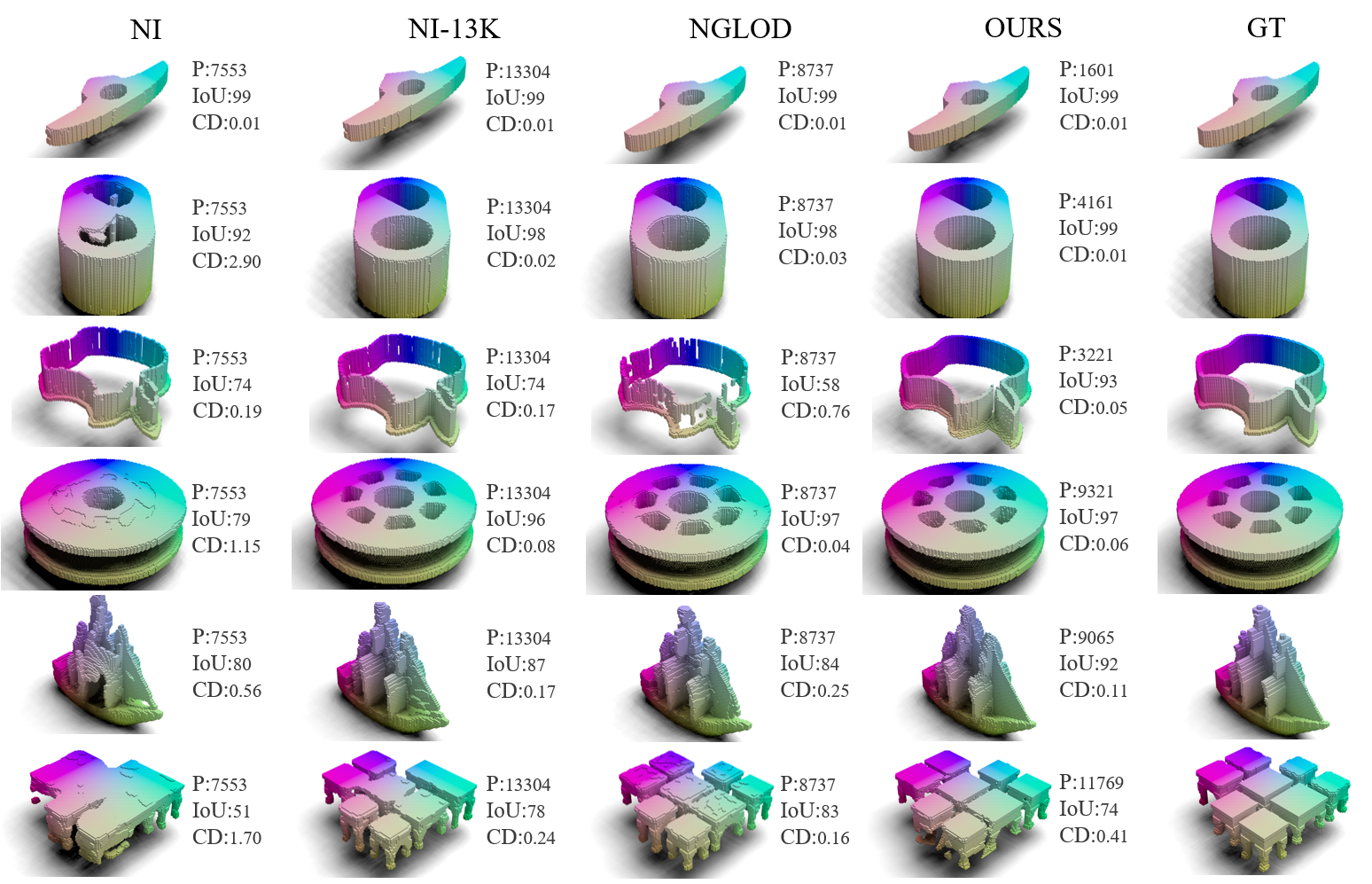}
	\vspace{-3mm}
	\caption{Comparison of reconstructed voxel models at $128^3$ resolution for different methods. All six models come from the Thingi10K dataset~\cite{Thingi10k}. The rendering method~\cite{Mitsuba2} is used for the colorful display. Two metrics IoU and CD are defined in Sec.~\ref{sec:4.1}. The number of network parameters is denoted by P. NI~\cite{NI} uses an MLP network, which by default has 8 hidden layers, each with 32 nodes. The enhanced NI-13K uses an MLP network with 8 hidden layers and 42 nodes per layer. NGLOD~\cite{nglod} stores 4737 network parameters and 32-dimensional latent vectors of 125 grid points.
	}
	\label{fig:reconstruction_comparison}
\end{figure*}

Faced with the success of SDF prediction networks, the first problem is why all existing high-precision reconstruction methods predict SDF.
Theoretically, it is much harder for a neural network to predict the SDF value of each point in the space than to predict its occupancy.
However, to the best of our knowledge, there is no neural network proposed to classify binary voxels for high-fidelity single-object reconstruction.
Predicting the occupancy of binary voxels with the same MLP network should yield higher reconstruction accuracy and fewer failures than predicting SDF values.

The second problem is that in MLP networks for model reconstruction, the number of layers, the number of neurons in each layer, and the activation functions are all set to be fixed.
However, the complexity of each model is different. 
3D models may be convex or concave, with or without holes.
In this sense, the number of layers and the number of neurons in each layer used to correctly reconstruct objects should be different.
On the other hand, activation functions tend to behave differently in different datasets~\cite{swish}. 
For the task of model reconstruction, each object is equivalent to a dataset with different shapes and data distributions. 
Naturally, for an individual object, the activation function of each layer selected by learning will be more suitable for model reconstruction.

In this paper, we propose an efficient end-to-end 3D voxel reconstruction based on neural architecture search (NAS)~\cite{nasrl}~\cite{dnasrl}.
NAS can find a specific network for an individual 3D object in terms of the number of layers, the number of nodes in each layer, and the activation function in each layer.
Moreover, directly predicting the occupancy of voxels not only alleviates the fitting difficulty for complex objects, but also avoids the use of surface reconstruction algorithms after obtaining SDF values.
The solution of the two problems mentioned above brings about a significant improvement in the reconstruction accuracy.
The comparison of six reconstructed voxel models for four methods with their ground truth (GT) is shown in Fig.~\ref{fig:reconstruction_comparison}.
The voxels in the $1$-st row is the simplest of the six, and all four methods achieve roughly the same reconstruction accuracy.
The voxels in the $2$-nd and $4$-th rows become more complex, and NI fails to reconstruct them correctly.
While the voxels in the $3$-rd row is not very complicated in geometric shape, only our result is visually acceptable.
For the voxels in the $5$-th row, the bow details of the boat can only be correctly reconstructed by our method.
The voxels in the $6$-th row is the most complex.
For our result, although the legs of the stools are not reconstructed well, there are no noticeable errors in the arresting surface of several stools in all results.
Moreover, only the proposed method has the adaptive number of network parameters for different objects, realizing the idea that simpler objects should have fewer parameters of neural representation.

Our contributions are summarized as follows:

1) MLP is used to directly predict the occupancy of each voxel, which significantly improves the reconstruction accuracy. Meanwhile, this end-to-end approach avoids the surface reconstruction required by SDF prediction methods.

2) The network architecture search (NAS) technology is used to find a specific network architecture for each object. 
The number of the network parameters can vary adaptively with the complexity of the object.

3) The network size is added to the reward, and a post-processing step after NAS is designed. By doing so, the number of network parameters is further reduced, while maintaining almost the same accuracy.

\vspace{-2mm}
\section{Related Works}
\vspace{-1mm}
This paper is mainly related to two directions, which are the neural implicit reconstruction of 3D models and the technology of network architecture search. They will be illustrated in the following two subsections, respectively.

\subsection{Neural Implicit Reconstruction}
With the development of 3D deep learning, there is a growing body of work studying implicit neural representation and reconstruction.
Here we investigate two sub-directions closely related to our work.
The first sub-direction in implicit neural representation and reconstruction is the prediction of SDFs from dense samples using MLP networks~\cite{deepsdf}~\cite{FFN}~\cite{SIREN}~\cite{NI}~\cite{nglod}~\cite{deepLS}~\cite{sn-graph}.
For example, DeepSDF~\cite{deepsdf} early proposes to learn and reconstruct continuous SDFs for a category of 3D objects using an MLP network.
FFN~\cite{FFN} maps Fourier features and learns high-frequency functions in low-dimensional domains to improve the fitting capability of MLP. 
To overcome the difficulty of fitting SDFs of one category of objects, NI~\cite{NI} firstly proposes to overfit the global shape of an individual object with a small MLP network. 
This method actually implements a lossy compression of a 3D model by storing MLP parameters.
However, the default network with 7553 parameters may fail in reconstruction, especially for complex objects.
NGLOD~\cite{nglod} learns the latent vectors of octree vertices in different levels of details (LOD) to predict local SDFs of an object.
Although the reconstruction accuracy can be improved obviously, the storage capacity in this local-fitting method is greatly increased as the latent vectors of a large number of LOD vertices needs to be stored.

The second sub-direction in implicit neural reconstruction is the prediction of occupancy of voxels using encoder-decoder networks~\cite{ONET}~\cite{learningimp}~\cite{convolutional}.
Similar to ONet~\cite{ONET} reviewed in Sec.~\ref{sec:1}, IM-Net~\cite{learningimp} learns the  generative models of shapes for one category of objects.
CONet~\cite{convolutional} combines convolutional encoders with implicit occupancy decoders to represent detailed reconstruction of objects and 3D scenes.
However, the reconstruction accuracy of these works is obviously lower than the SDF prediction networks.
Moreover, the network size of these works is much larger than most of MLP networks.

In theory, predicting occupancy of points is much easier than predicting their SDF values with the same network and training data.
Therefore, in this paper, we adopt this idea to directly predict the occupancy of binary voxels.

\subsection{network architecture search}
Neural Architecture Search (NAS) methods essentially aim to provide an automated way to design architectures as an alternative to manual architectures. 
Our work is closely related to reinforcement learning based NAS work~\cite{nasrl}~\cite{dnasrl}~\cite{SIR}~\cite{nasreg}~\cite{ENAS}. 
For example, owing to the weight sharing idea, ENAS~\cite{ENAS} can significantly reduce the computational power required to traverse the search space~\cite{SIR}. 
Although later researchers propose a different framework DARTS ~\cite{darts}~\cite{darts+}~\cite{idarts}, this kind of approach is not suitable for MLP architectures search. 

In addition, the process of NAS needs to accurately evaluate the performance of each network architecture. 
A straightforward solution is to train an architecture from scratch and then test it on the validation dataset, which is very time consuming. 
Instead of accurately evaluating the network architecture on the target task, the researchers proposed the proxy task method. 
The proxy task means training on subsets of dataset or fine-tuning with fewer epochs~\cite{ENAS}~\cite{mobjnas}~\cite{bayesiannas}~\cite{fbnet}. 
Although these methods improve the speed of NAS, a rough evaluation inevitably treats some promising network architectures as poor networks. A post-processing step after NAS is proposed in this paper.
As a result, those potential networks with fewer network parameters can be found.

\section{The Proposed Method}
The purpose of this paper is to utilize NAS technology to search for specific network architectures for different individual objects, while completing end-to-end neural reconstruction through binary classification of voxels.
Compared with previous works~\cite{ONET}~\cite{NI}~\cite{nglod}, this adaptive reconstruction method achieves higher reconstruction accuracy using fewer network parameters.
The following subsections describe the proposed method in detail.

\subsection{Binary Voxel Classification}
Binary voxels can be directly visualized as one of the explicit representations of 3D objects.
In general, voxels inside an object are defined as 1, while voxels outside the object are defined as 0.
Training a neural network to classify the binary voxels of a given model in 3D space enables end-to-end neural representation and reconstruction.
Therefore, the post-processing steps, such as surface reconstruction using marching cubes~\cite{NI}~\cite{nglod}, can be avoided.

Objects are normalized in a 3D space denoted by $\mathcal{H}\!=\![-1,1]^3$.
The entire normalized space is divided equally into $N^3$ parts, each of which is regarded as a voxel.
The set of voxels inside the object is denoted by $\mathcal{V}$.
For each voxel $\mathbf{p}$ in $\mathcal{H}$, the neural network $f_{\theta}(\cdot)$ outputs the probability of $\mathbf{p} \in \mathcal{V}$, which is between 0 and 1.
To optimize the parameters $\theta$ of the neural network, the cross-entropy classification loss $\mathcal{L}(\theta)$ is used as the following,
\vspace{-3mm}

\begin{small}
\begin{equation}
\begin{split}
    &\mathcal{L}(\theta)\!=\!\frac{1}{K}\!\sum_{i=1}^K-[y_{i}\log(f_{\theta}(\mathbf{p}_{i})) + (1\!-\!y_{i})\log(1\!-\!f_{\theta}(\mathbf{p}_{i}))], \\
    &\text{with} \quad y_{i}=\left\{
        \begin{array}{lr}
            1 \quad if \ \mathbf{p}_{i}\in \mathcal{V},\\
            0  \quad otherwise.
        \end{array}
    \right. 
\end{split}
\end{equation}
\end{small}

\vspace{-1mm}
\noindent where $\mathbf{p}_{i}$ is the $i$-th voxel in $K$ sampled voxels, which will be explained in Sec.~\ref{sec:3-5}.

\subsection{Search Space of Neural Architecture}
Most of previous works employ MLP as their global or local SDF fitting network.
Therefore, the number of layers, the number of neurons in each layer, and the activation functions in each layer are chosen as our neural architecture search space.

Specifically, in order to reduce the search time and take into account the reasonable search range, the search space for the number of nodes is \{8, 12, 16, 20, 24, 28, 32, 40, 48, 56, 64\}.
Moreover, we let the controller decide the activation function for each layer, making the network architecture more expressive. 
The search space of activation functions is
\{$ReLU, ELU, Swish$\}~\cite{relu}~\cite{elu}~\cite{swish}, each of which can be represented by 
\vspace{-4mm}

\begin{small}
\begin{equation}
\begin{split}
        ReLU&:g(x)  = \begin{cases}
      x& \text{ if } x\ge  0 \\
      0& \text{ if } x< 0
    \end{cases}
    \\
    ELU&:g(x)  = \begin{cases}
      x & \text{ if } x\ge 0 \\
      \alpha \left ( \exp\left ( x \right ) -1 \right ) & \text{ if } x< 0
    \end{cases}
    \\
    Swish&:g(x)=x\cdot Sigmoid(\beta x)
\end{split}
\end{equation}
\end{small}

\vspace{-1mm}
\noindent where $\alpha$ and $\beta$ are usually set to 1.

In the above selection of activation functions, the traditional activation functions $Sigmoid$ and $Tanh$ are not included.
This is because we experimentally find that adding these activation functions will reduce the classification performance of the network.
As stated in the previous NAS works~\cite{MnasNet}~\cite{RLNas}~\cite{SIR}, the choice and design of the search space plays a crucial role in the success of NAS. 
Ablation experiments shown in Sec.~\ref{sec:4.4} demonstrate the performance of adding activation functions into the search space. 
\begin{figure}[t]
	\centering
	\includegraphics[width=0.5\textwidth]{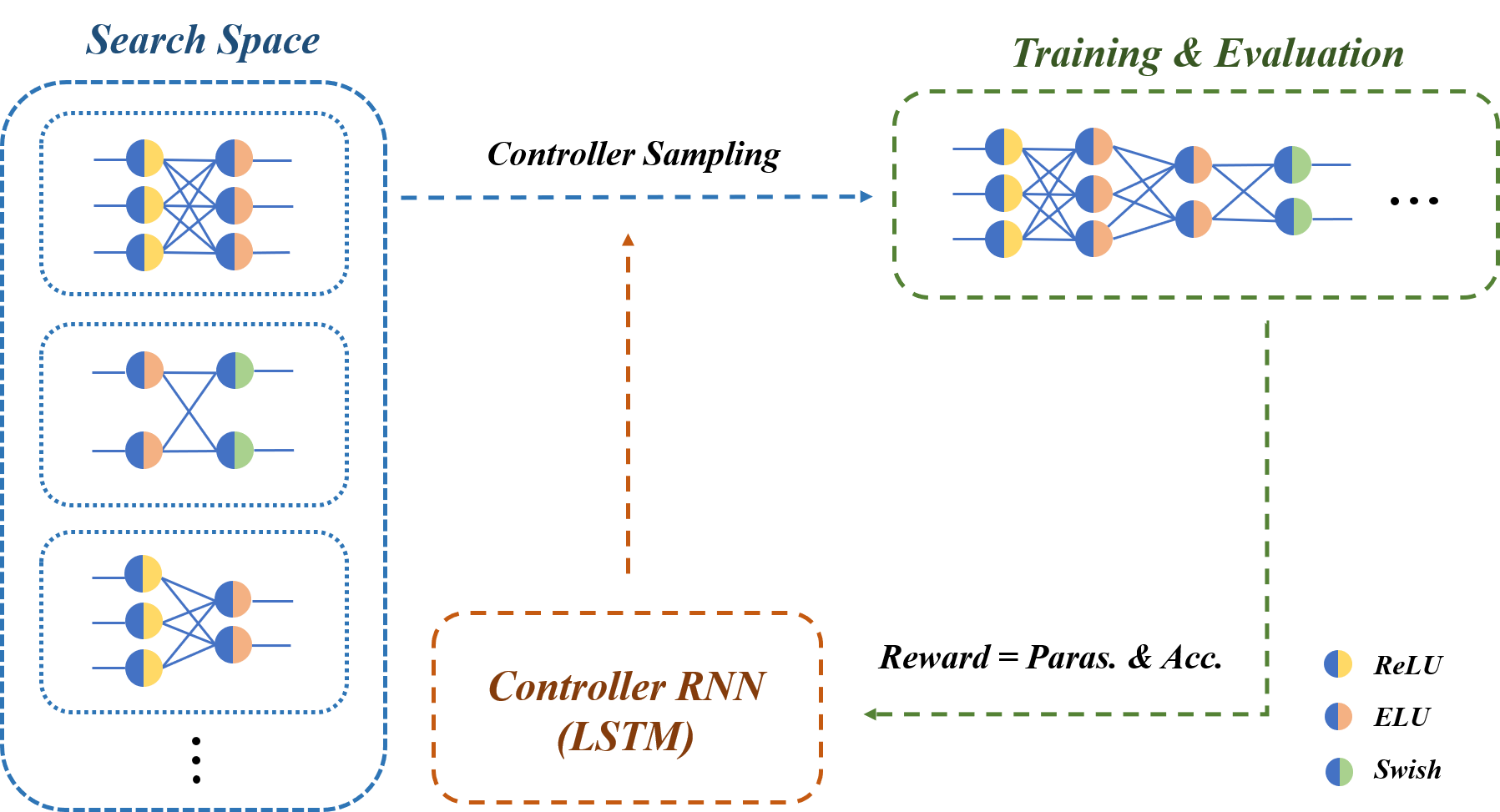}
	\vspace{-4mm}
	\caption{Searching process of neural architecture.}
	\label{fig:searching_process}
	\vspace{-3mm}
\end{figure}

\subsection{Process of NAS}
In order to search a `proper' neural architecture specifically for a given object, we utilize a mature NAS algorithm ENAS~\cite{ENAS}. 
The searching process is drawn in Fig.~\ref{fig:searching_process}.
The controller samples the MLP layers from the search space consisting of different numbers of nodes and activation functions.
After an MLP network is trained, its reward can be obtained on the validation set.
The reward is then fed back to the controller RNN for policy-gradient descent.

Unlike ENAS, the search strategy in our method focuses on how to choose a better MLP network with different numbers of neurons and activation functions.
However, the raw reward in ENAS only considers the classification score, and the internal controller always generates network architectures that make the classification reward higher.
Since one of the expectations of the neural reconstruction is to minimize the number of network parameters~\cite{NI}, the impact of network size should be factored into the reward. 
Thus, the reward is designed as 
\vspace{-3mm}

\begin{small}
\begin{equation}
reward=(Acc_{val}\!-\!Acc_{base})+(P_{base}\!-\!P_{val})/P_{max}
\label{equ:1}
\end{equation}
\end{small}

\vspace{-1mm}
\noindent where $Acc_{val}$ denotes the classification accuracy of the evaluated network architecture for all voxels. 
$Acc_{base}$ is an expected accuracy, which is set to 0.98.
$P_{base}$ is equal to the parameter amount 7553 of the default network of NI.
$P_{val}$ denotes the size of the evaluated network architecture.
$P_{max}$ is also a fixed value 21121, which is the number of the largest network architecture parameter in our search space.

It is worth noting that if the controller generates an output layer in the process of generating MLP layers, the number of MLP layers will no longer increase.
This explains that the number of MLP layers can also be searched during the NAS process.

\subsection{Post-processing Step after NAS}
After completing the NAS process, we introduce a post-processing step to select smaller neural architectures.
There are two reasons for this step.
First, to speed up the NAS process, we use a proxy task~\cite{SIR} that is similar to NAS-FPN~\cite{NAS-FPN} to shorten the training time of the target task.
We train the proxy task $e_1$ epochs instead of $e_2$ epochs used to train the target network.
This early termination method speeds up the convergence time of rewards by a factor of $e_1/e_2$.
Since the proxy task does not train the network to converge, the ranking of the network accuracy cannot represent the final ranking of the network.
Therefore, the network with accuracy slightly lower than the highest accuracy can be considered as candidates.

Second, although we have used the network size reward term in Eq.~\ref{equ:1} to influence the network chosen by NAS, a fixed weight is not applicable to different objects.
A network with a little lower network size and classification accuracy may be excluded, compared to the network with the least reward in the NAS process.
Post-processing selection of multiple candidates would greatly alleviate this problem.

Specifically, we filter out all candidate networks whose accuracy is lower than the highest accuracy in the NAS process up to a threshold $t$ ($t\!=\!0.1\%$ in experiments).
Then the network with the smallest size will be selected.
The ablation experiments shown in Sec.~\ref{sec:4.3} validate the effectiveness of the proposed post-processing step.

\subsection{Other Details of Data Processing, Sampling, and NAS Configuration}
\label{sec:3-5}
In data processing, 3D models are firstly normalized in a sphere with the radius $0.9$. 
Then we utilize \textit{PyMesh} library to voxelize a model to $N^3$ resolution (N=128 by default).

In voxel sampling, we first sample all surface voxels and their outer layer voxels as the support samples of classification boundary.
Then, we down-sample $1/4$ other non-support voxels and copy the support voxels to the same number.
Thus, the total sample number $K$ is $1/2$ the number of other voxels, which is roughly equivalent to 1M. 

In the NAS configuration, the maximum number of total network layers and hidden layers is 8 and 6, respectively. 
The controller samples 6 MLP layers of different architectures at a time.
Due to the weight sharing strategy proposed in ENAS, we terminate the search by only sampling 5 times.

\section{Experiments}
Various experiments are conducted to verify reconstruction quality of the proposed end-to-end method.
Sec.~\ref{sec:4.1} describes the used datasets and metrics.
Sec.~\ref{sec:4.2} shows the experimental results, compared with other methods. 
Sec.~\ref{sec:4.3} gives the first ablation experiment of removing NAS and other improvements.
Sec.~\ref{sec:4.4} gives the second ablation experiment of activation functions.
\textbf{The pre-trained network models for all displayed objects can be reproduced with our source code in} \href{https://github.com/cscvlab/VoxelReconstruction-NAS}{https://github.com/cscvlab/VoxelReconstruction-NAS}.

\vspace{-1mm}
\subsection{Datasets and Metrics}
\vspace{-1mm}
\label{sec:4.1}
Datasets we used in this paper include Thingi10K~\cite{Thingi10k}, Thingi32 and ShapeNet150. 
Thingi10K is composed of $10,000$ 3D-printing
models, which have been tested in NI~\cite{NI}.
NGLOD~\cite{nglod} mainly verifies two other small datasets: Thingi32 and ShapeNet150. 
Thingi32 contains 32 simple shapes in Thingi10K. ShapeNet150 contains 150 shapes in the ShapeNet dataset~\cite{shapenet}, including 50 cars, 50 airplanes, and 50 chairs. 

The metrics for evaluation are common  3D intersection over union (IoU) and Chamfer distance (CD). 
The former metric 3D IoU is defined as the ratio of the intersection and the union voxels of a reconstructed model and its ground truth model.
The latter metric CD is defined as the bi-directional minimum distance~\cite{pytorch3D} from the surface voxels $\mathcal{S}_{r}$ of a reconstructed model to the surface voxels $\mathcal{S}_{r}$ of its ground truth model, which is expressed by
\vspace{-3mm}

\begin{small}
\begin{equation}
    \mathrm{CD}(\mathcal{S}_{r},\mathcal{S}_{g})
    =\frac{1}{n_{r}}\!\sum_{\mathbf{p}_{r} }\min_{\mathbf{p}_{g}}\|\mathbf{p}_{r}\!-\!\mathbf{p}_{g}\|_{2}^{2}+\frac{1}{n_{g}}\!\sum_{\mathbf{p}_{g}}\min_{\mathbf{p}_{r}}\|\mathbf{p}_{g}\!-\!\mathbf{p}_{r}\|_{2}^{2}
\end{equation}
\end{small}

\vspace{-2mm}
\noindent where $n_r$ and $n_g$ denote the voxel number of $\mathcal{S}_{r}$ and $\mathcal{S}_{g}$, respectively. $\mathbf{p}_r$ and $\mathbf{p}_g$ denote each voxel in $\mathcal{S}_{r}$ and $\mathcal{S}_{g}$, respectively.
In all the following results, the value of CD is magnified by a factor of 1000 for convenience of display.

\vspace{-1mm}
\subsection{Comparison with Previous Methods}
\label{sec:4.2}
\vspace{-1mm}
We compare our approach with one voxel reconstruction method ONet~\cite{ONET}, and two SDF prediction methods NI~\cite{NI} (with the default configuration) and NGLOD~\cite{nglod} (with LOD level 1).
First, the comparison is conducted on Thingi32 and ShapeNet150 datasets.
The experimental results of the previous three methods are shown in the upper part of Table~\ref{table:1}, and our results are shown in the last row.
Note that the results of NI-Thingi32 and ONet-ShapeNet150 are obtained through the official trained networks.
The results of NI-ShapeNet150 and NGLOD on two datasets are obtained by our training using the official codes.
Since ONet is trained on one category of objects, it cannot be used for objects in Thingi32.
\begin{table}[tb!]
\begin{center}
\caption{Comparison with previous methods and three ablation ways.}
\label{table:1} 
    \vspace{-2mm}
    \resizebox{0.48\textwidth}{!}
    {
    \begin{tabular}{|l||c|c|c|c|c|c|}
    \hline
    \multirow{2}{*}{Method}  & 
    \multicolumn{3}{c|}{Thingi32} & \multicolumn{3}{c|}{ShapeNet150} \\
    \cline{2-7}
    & Size & IoU & CD & Size & IoU & CD 
    \\
    \hline
    ONet~\cite{ONET}  &--  &--  &--  &6M  & 60.7 & 4.613 \\ \hline 
    NI~\cite{NI}  & 7553 & 96.8 & 0.117 & 7553 & 82.7 & 4.326 \\ \hline
    NGLOD~\cite{nglod}  &8737  &97.4  &0.088  &8737  &82.5  &1.163  \\ \hline 
    \hline
    \hline
    Ours w.o. NAS  & 7553 & 97.8 & 0.089 & 7553 & 95.5 & 0.084 \\ \hline
        \tabincell{c}{Ours w. NAS\\ (w.o. RI\&PPS)} & 8837 & \textbf{98.0} & \textbf{0.076} & 8551 & \textbf{96.0} & \textbf{0.071} \\ \hline
    \tabincell{c}{Ours w. NAS\\ (w.o. PPS)} & 7626 & 97.2 & 0.110 & 7461 & 95.0 & 0.103 \\ \hline
    Ours w. NAS & \textbf{5452} & 97.4 & 0.100 & \textbf{5860} & 95.7 & 0.082 \\ \hline
    \end{tabular}
    }
    \vspace{-6mm}
    \end{center}
    \end{table}

\begin{figure*}[t]
	\centering
	\includegraphics[width=0.92\textwidth]{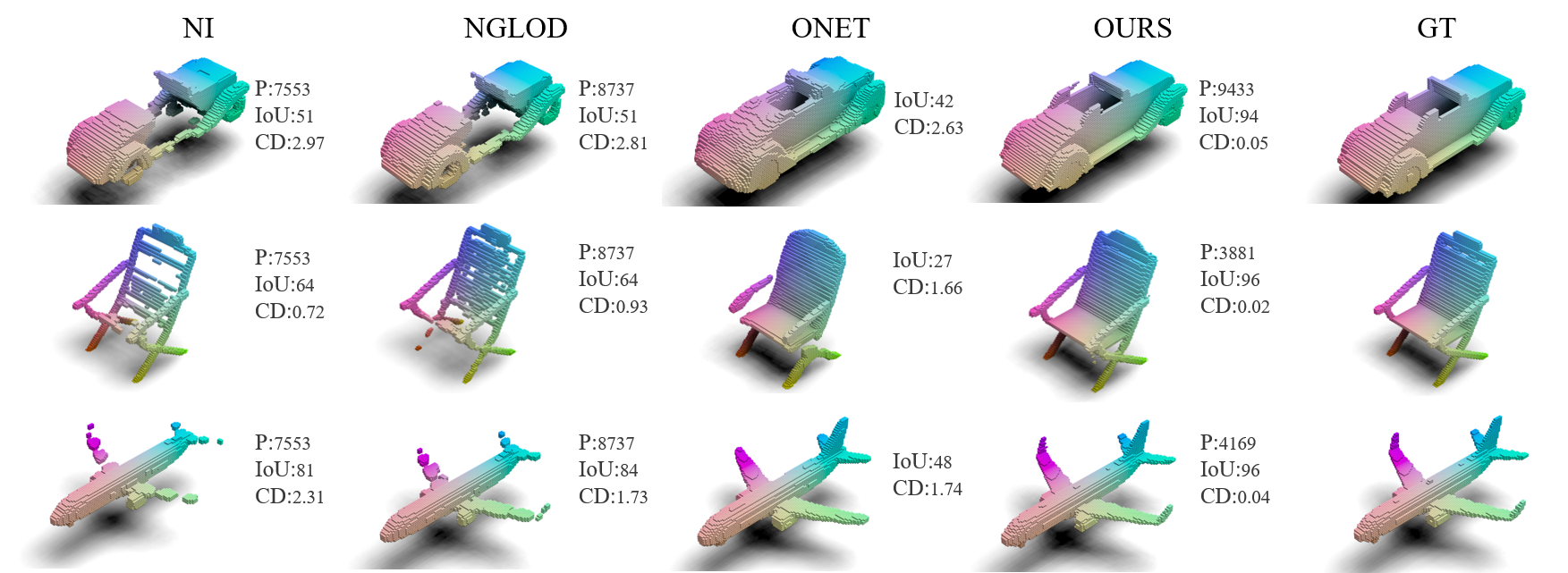}
	\vspace{-2mm}
	\caption{Visual comparison of three reconstructed voxel models (one for each category in ShapeNet150) for the four methods.}
	\label{fig:shapenet_comparison}
	\vspace{-6mm}
\end{figure*}

Since Thingi32 only contains models of simple shapes, NI, NGLOD and our method perform similarly, and they can all reconstruct these models without noticeable errors. 
For ShapeNet150, the performance differences of the four methods are easily distinguishable.
Since there are some complicated objects in ShapeNet150, ONet and NI cannot reconstruct them correctly.
This results in a large increase in the metric CD.
NGLOD can handle more objects correctly than ONet and NI, and gain better CD and IoU.
The proposed method shows significant improvement in CD and IoU, which indicates that our method does not suffer as much performance degradation as other methods when dealing with complex objects.
Fig.~\ref{fig:shapenet_comparison} depicts three reconstructed voxels (one for each category in ShapeNet150) for all four methods as a visual comparison.

\begin{figure}[thb!]
\centering
\begin{minipage}[b]{0.24\textwidth}
\centering
\includegraphics[width=1\textwidth]{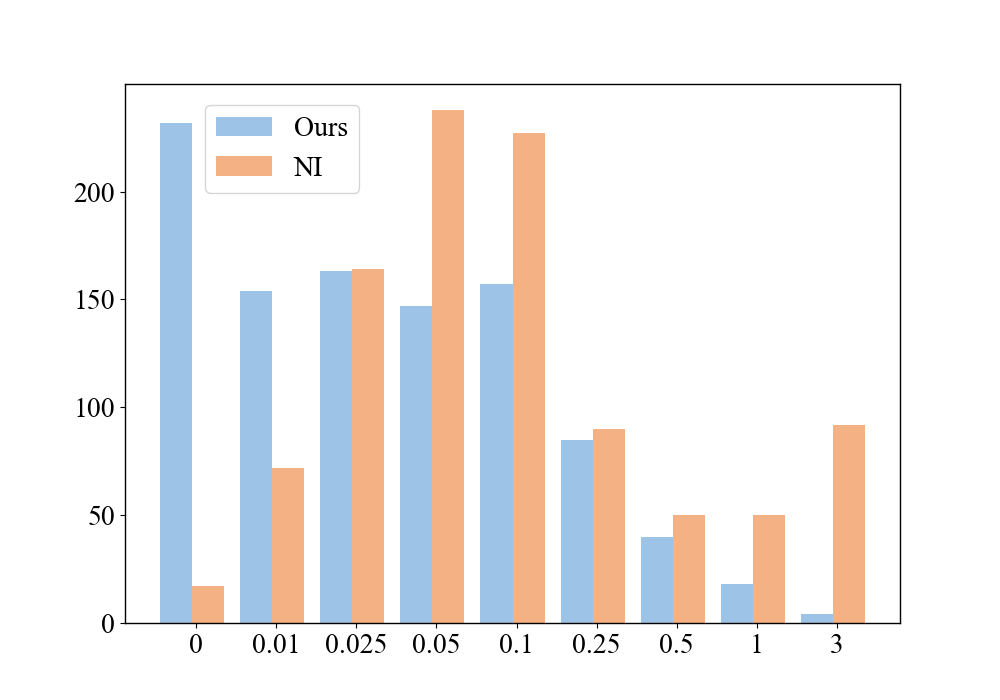}
\end{minipage}
\begin{minipage}[b]{0.24\textwidth}
\includegraphics[width=1\textwidth]{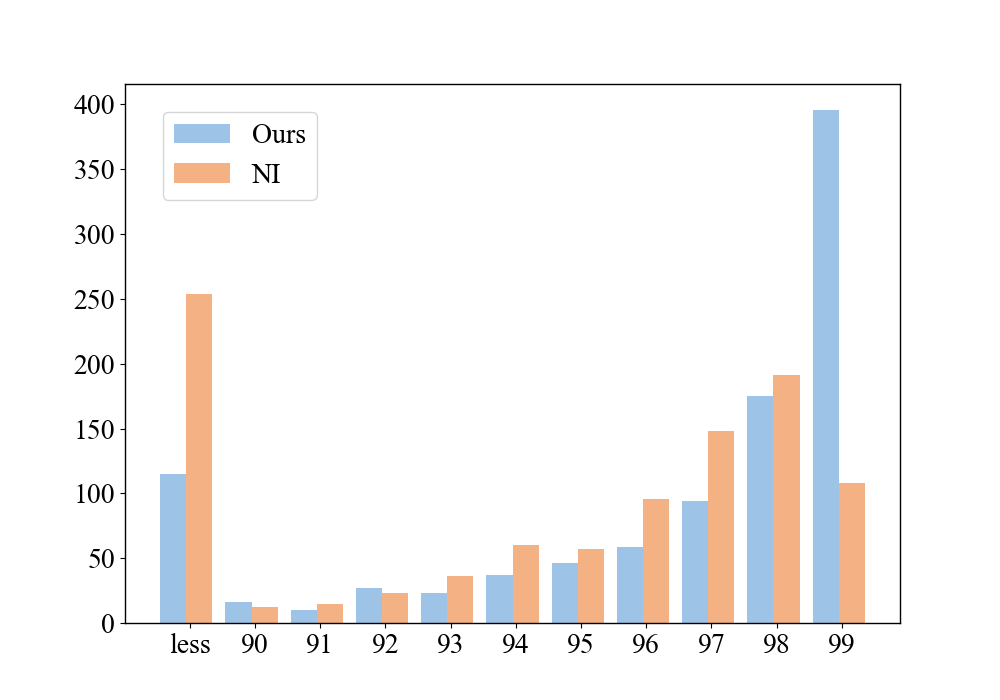}
\end{minipage}
\vspace{-8mm}
\caption{Histograms of CD (on the left) and IoU (on the right) on 1000 randomly selected objects in Thingi10K.}
\label{fig:thingi1K_comparison}
\vspace{-4mm}
\end{figure}


As ShapeNet150 contains only 150 objects in three categories, we further conduct another experiment on Thingi10K dataset. 
We train the NAS networks for $1,000$ models randomly selected from Thingi10K, and compare the reconstruction results with NI.
Fig.~\ref{fig:thingi1K_comparison} depicts the histograms of CD and IoU for the two methods, respectively.
The proposed method is obviously superior to NI.

\vspace{-1mm}
\subsection{Ablation Experiments of Removing NAS, Size Reward and Post-processing Step}
\label{sec:4.3}
\vspace{-1mm}
This ablation experiment aims to observe the influence of removing NAS, the proposed size reward, and the post-processing step.
The experimental results are shown in the lower part of Table~\ref{table:1} (rows $5$ to $7$).
The $5$-th row shows our results without NAS.
The network here is much the same as NI. 
The main difference is that instead of predicting SDF values, we directly classify voxels, and over-sample the support voxels.
As NI successfully reconstructs the 3D models in Thingi32, the method in the $5$-th row achieves only a slight improvement.
However, for the more complicated ShapeNet150 dataset, the improvements of IoU and CD are significant. 
This means that binary classification is much easier to be fitted by a same network than SDF prediction, which experimentally validates the first theoretical problem described in Sec.~\ref{sec:1}.

The $6$-th row shows our results using NAS, but without adding network size to reward and the post-processing step.
Since there is no scheme to control the network scale, the NAS obtains networks with more parameters at average.   
Owing to larger network parameters, the method in the $6$-th row gains the best IoU and CD performance on two datasets.

The $7$-th row shows our results using NAS and size reward, but without the post-processing step.
Since the reward is improved to control network size, the mean number of network parameters is roughly the same to it in the $5$-th row, and is reduced by $\sim$$15\%$ compared to the $6$-th row.
As a result, the performance of the method in the $7$-th row degrades slightly.

The advantage of the post-processing step is reflected in the last row. 
Compared to the $7$-th row, although the performance is promoted slightly, the mean size of the networks is significantly reduced.
This verifies the effectiveness of the proposed method in selecting a suitable network architecture.

\vspace{-2mm}
\subsection{Ablation Experiment of Activation Functions}
\label{sec:4.4}
\vspace{-1mm}
In order to choose suitable activation functions as the candidates in our search space, we test five activation functions, which are Sigmoid, Tanh, ReLU, ELU, and Swish, on Thingi32 and ShapeNet150 separately.
This is also an ablation experiment, as NAS only searches different numbers of layers and nodes at the moment. 
There is no change to the setting except that only one activation function is used during NAS.
The experimental results are shown in Table~\ref{table:2}.

The activation function ReLU, ELU or Swish, can achieve good accuracy, while the results of Sigmoid are dramatically worse, especially on ShapeNet150.
Although the mean network size of Sigmoid appears to be minimal, we experimentally find that the reconstruction for Sigmoid is likely to fail.
In order to reduce invalid searches, we remove Sigmoid and Tanh from the final search space of activation functions.
Compared to using one fixed activation function, using three activation functions \{ReLU, ELU, Swish\} as the search space reduces CD and IoU very slightly, but searches obviously smaller networks. 
\begin{table}[tb!]
\begin{center}
\caption{Ablation experiment of activation functions.}
\label{table:2} 
    \vspace{-1mm}
    \resizebox{0.48\textwidth}{!}
    {
    \begin{tabular}{|l||c|c|c|c|c|c|}
    \hline
    \multirow{2}{*}{Method}  & 
    \multicolumn{3}{c|}{Thingi32} & \multicolumn{3}{c|}{ShapeNet150} \\
    \cline{2-7}
    & Size & IoU & CD & Size & IoU & CD 
    \\
    \hline
    Sigmoid & \textbf{4369} & 80.8 & 2.565 & \textbf{3217} & 58.3 & 18.420 \\ \hline
    Tanh & 5953 & 96.5 & 0.155 & 7098 & 94.9 & 0.125 \\ \hline 
    ReLU & 6455 & \textbf{98.2} & \textbf{0.071} &  6040 & 95.4 & 0.132 \\ \hline
    ELU & 6149 & 96.7 & 0.147 & 6697 & 94.6 & 0.110  \\ \hline 
    Swish & 5886 & 97.4 & 0.106 & 6236 & 95.6 & 0.098 \\ \hline
    \tabincell{c}{\{ReLU,\\ELU,\\Swish\}} & 5452 & 97.4 & 0.100 & 5860 & \textbf{95.7} & \textbf{0.082}\\ \hline
    \end{tabular}
    }
    \vspace{-4mm}
    \end{center}
    \end{table}

\section{Conclusion}
This paper proposes a neural implicit reconstruction method of 3D objects based on network architecture search (NAS). 
Without any surface reconstruction algorithm (e.g., marching cube~\cite{marching}), we employ an end-to-end network by directly classifying binary voxels.
Although the basic idea is straightforward to some extent, the proposed approach outperforms the state-of-the-art methods~\cite{NI}~\cite{nglod} using SDF prediction network and the marching cube algorithm.
From various conducted experiments, we can conclude that the combination of different layers, node numbers, activation functions (searched by NAS), and using binary classification together lead to the improvement of reconstruction quality, especially at classification boundaries.
Furthermore, the number of network parameters is added to the reward during NAS, which reduces the storage of the neural implicit representation.
In other words, the further improvement of the compression ratio enhances the storage advantage of neural implicit representation over traditional explicit representations.

One disadvantage of the proposed method is that its flexibility is temporarily limited, since the learned discrete voxels have a fixed resolution. 
For continuous SDF prediction networks, voxel models at any resolution can be generated by the marching cube algorithm without re-training the networks.
This problem may be addressed by future work incorporating the octree representation of voxel, which can progressively classify more subdivided voxels.

\section*{Acknowledgment}
This work is supported by Natural Science Foundation of Shanghai (Grant No. 21ZR1401200), Shanghai Sailing Program (21YF1401300), and the Foundation of Key Laboratory of Artificial Intelligence, Ministry of Education, P.R. China (AI2020003).

\bibliographystyle{IEEEtran}
\bibliography{ref.bib}

\begin{thebibliography}{10}
\providecommand{\url}[1]{#1}
\csname url@samestyle\endcsname
\providecommand{\newblock}{\relax}
\providecommand{\bibinfo}[2]{#2}
\providecommand{\BIBentrySTDinterwordspacing}{\spaceskip=0pt\relax}
\providecommand{\BIBentryALTinterwordstretchfactor}{4}
\providecommand{\BIBentryALTinterwordspacing}{\spaceskip=\fontdimen2\font plus
\BIBentryALTinterwordstretchfactor\fontdimen3\font minus
  \fontdimen4\font\relax}
\providecommand{\BIBforeignlanguage}[2]{{%
\expandafter\ifx\csname l@#1\endcsname\relax
\typeout{** WARNING: IEEEtran.bst: No hyphenation pattern has been}%
\typeout{** loaded for the language `#1'. Using the pattern for}%
\typeout{** the default language instead.}%
\else
\language=\csname l@#1\endcsname
\fi
#2}}
\providecommand{\BIBdecl}{\relax}
\BIBdecl

\bibitem{pointnet}
R.~Q. Charles, H.~Su, M.~Kaichun, and L.~J. Guibas, ``Pointnet: Deep learning
  on point sets for 3d classification and segmentation,'' in \emph{IEEE/CVF
  Conference on Computer Vision and Pattern Recognition (CVPR)}, 2017, pp.
  77--85.

\bibitem{voxnet}
D.~Maturana and S.~Scherer, ``Voxnet: A 3d convolutional neural network for
  real-time object recognition,'' in \emph{IEEE/RSJ International Conference on
  Intelligent Robots and Systems (IROS)}, 2015, pp. 922--928.

\bibitem{ONET}
L.~Mescheder, M.~Oechsle, M.~Niemeyer, S.~Nowozin, and A.~Geiger, ``Occupancy
  networks: Learning 3d reconstruction in function space,'' in \emph{IEEE/CVF
  Conference on Computer Vision and Pattern Recognition (CVPR)}, 2019, pp.
  4455--4465.

\bibitem{NI}
T.~Davies, D.~Nowrouzezahrai, and A.~Jacobson, ``On the effectiveness of
  weight-encoded neural implicit 3d shapes,'' \emph{arXiv preprint
  arXiv:2009.09808}, 2020.

\bibitem{nglod}
T.~Takikawa, J.~Litalien, K.~Yin, K.~Kreis, C.~Loop, D.~Nowrouzezahrai,
  A.~Jacobson, M.~McGuire, and S.~Fidler, ``Neural geometric level of detail:
  Real-time rendering with implicit 3d shapes,'' in \emph{IEEE/CVF Conference
  on Computer Vision and Pattern Recognition (CVPR)}, 2021, pp.
  11\,353--11\,362.

\bibitem{nerf}
B.~Mildenhall, P.~P. Srinivasan, M.~Tancik, J.~T. Barron, R.~Ramamoorthi, and
  R.~Ng, ``Nerf: Representing scenes as neural radiance fields for view
  synthesis,'' in \emph{European conference on computer vision (ECCV)}, 2020,
  pp. 405--421.

\bibitem{nerf++}
M.~Niemeyer and A.~Geiger, ``Giraffe: Representing scenes as compositional
  generative neural feature fields,'' in \emph{IEEE/CVF Conference on Computer
  Vision and Pattern Recognition (CVPR)}, 2021, pp. 11\,448--11\,459.

\bibitem{qiuxieNerf}
S.~Fridovich-Keil, A.~Yu, M.~Tancik, Q.~Chen, B.~Recht, and A.~Kanazawa,
  ``Plenoxels: Radiance fields without neural networks,'' in \emph{Proceedings
  of the IEEE/CVF Conference on Computer Vision and Pattern Recognition
  (CVPR)}, 2022, pp. 5501--5510.

\bibitem{Thingi10k}
Q.~Zhou and A.~Jacobson, ``Thingi10k: A dataset of 10,000 3d-printing models,''
  \emph{arXiv preprint arXiv:1605.04797}, 2016.

\bibitem{Mitsuba2}
M.~Nimier-David, D.~Vicini, T.~Zeltner, and W.~Jakob, ``Mitsuba 2: a
  retargetable forward and inverse renderer,'' \emph{ACM Transactions on
  Graphics}, vol.~38, pp. 1--17, 2019.

\bibitem{swish}
P.~Ramachandran, B.~Zoph, and Q.~V. Le, ``Searching for activation functions,''
  \emph{arXiv preprint arXiv:1710.05941}, 2017.

\bibitem{nasrl}
B.~Zoph and Q.~V. Le, ``Neural architecture search with reinforcement
  learning,'' \emph{arXiv preprint arXiv:1611.01578}, 2016.

\bibitem{dnasrl}
B.~Baker, O.~Gupta, N.~Naik, and R.~Raskar, ``Designing neural network
  architectures using reinforcement learning,'' \emph{arXiv preprint
  arXiv:1611.02167}, 2016.

\bibitem{deepsdf}
J.~J. Park, P.~Florence, J.~Straub, R.~Newcombe, and S.~Lovegrove, ``Deepsdf:
  Learning continuous signed distance functions for shape representation,'' in
  \emph{IEEE/CVF Conference on Computer Vision and Pattern Recognition (CVPR)},
  2019, pp. 165--174.

\bibitem{FFN}
M.~Tancik, P.~Srinivasan, B.~Mildenhall, S.~Fridovich-Keil, N.~Raghavan,
  U.~Singhal, R.~Ramamoorthi, J.~Barron, and R.~Ng, ``Fourier features let
  networks learn high frequency functions in low dimensional domains,'' in
  \emph{Proceedings of the International Conference on Neural Information
  Processing Systems (NeurIPS)}, 2020, pp. 7537--7547.

\bibitem{SIREN}
V.~Sitzmann, J.~N. Martel, A.~W. Bergman, D.~B. Lindell, and G.~Wetzstein,
  ``Implicit neural representations with periodic activation functions,''
  \emph{arXiv preprint arXiv:2006.09661}, 2020.

\bibitem{deepLS}
R.~Chabra, J.~E. Lenssen, E.~Ilg, T.~Schmidt, J.~Straub, S.~Lovegrove, and
  R.~Newcombe, ``Deep local shapes: Learning local sdf priors for detailed 3d
  reconstruction,'' in \emph{European Conference on Computer Vision (ECCV)},
  2020, pp. 608--625.

\bibitem{sn-graph}
S.~Zhang, H.~Cao, Y.~Liu, S.~Cai, Y.~Zhang, Y.~Li, and X.~Chi, ``Sn-graph: A
  minimalist 3d object representation for classification,'' in \emph{IEEE
  International Conference on Multimedia and Expo (ICME)}, 2021, pp. 1--6.

\bibitem{learningimp}
Z.~Chen and H.~Zhang, ``Learning implicit fields for generative shape
  modeling,'' in \emph{IEEE/CVF Conference on Computer Vision and Pattern
  Recognition (CVPR)}, 2019, pp. 5932--5941.

\bibitem{convolutional}
S.~Peng, M.~Niemeyer, L.~Mescheder, M.~Pollefeys, and A.~Geiger,
  ``Convolutional occupancy networks,'' in \emph{European Conference on
  Computer Vision (ECCV)}, 2020, pp. 523--540.

\bibitem{SIR}
B.~Zoph, V.~Vasudevan, J.~Shlens, and Q.~V. Le, ``Learning transferable
  architectures for scalable image recognition,'' in \emph{IEEE/CVF Conference
  on Computer Vision and Pattern Recognition (CVPR)}, 2018, pp. 8697--8710.

\bibitem{nasreg}
E.~Real, A.~Aggarwal, Y.~Huang, and Q.~V. Le, ``Regularized evolution for image
  classifier architecture search,'' in \emph{Proceedings of the AAAI Conference
  on Artificial Intelligence (AAAI)}, 2019, pp. 4780--4789.

\bibitem{ENAS}
H.~Pham, M.~Y. Guan, B.~Zoph, Q.~V. Le, and J.~Dean, ``Efficient neural
  architecture search via parameter sharing,'' \emph{arXiv preprint arXiv:
  Learning}, 2018.

\bibitem{darts}
H.~Liu, K.~Simonyan, and Y.~Yang, ``Darts: Differentiable architecture
  search,'' \emph{arXiv preprint arXiv:1806.09055}, 2018.

\bibitem{darts+}
H.~Liang, S.~Zhang, J.~Sun, X.~He, W.~Huang, K.~Zhuang, and Z.~Li, ``Darts+:
  Improved differentiable architecture search with early stopping,''
  \emph{arXiv preprint arXiv:1909.06035}, 2019.

\bibitem{idarts}
H.~Wang, R.~Yang, D.~Huang, and Y.~Wang, ``idarts: Improving darts by node
  normalization and decorrelation discretization,'' \emph{IEEE Transactions on
  Neural Networks and Learning Systems, (online first)}, 2021.

\bibitem{mobjnas}
T.~Elsken, J.~H. Metzen, and F.~Hutter, ``Efficient multi-objective neural
  architecture search via lamarckian evolution,'' \emph{arXiv preprint
  arXiv:1804.09081}, 2018.

\bibitem{bayesiannas}
A.~Klein, S.~Falkner, S.~Bartels, P.~Hennig, and F.~Hutter, ``Fast bayesian
  hyperparameter optimization on large datasets,'' \emph{Electronic Journal of
  Statistics}, vol.~11, no.~2, pp. 4945--4968, 2017.

\bibitem{fbnet}
B.~Wu, X.~Dai, P.~Zhang, Y.~Wang, F.~Sun, Y.~Wu, Y.~Tian, P.~Vajda, Y.~Jia, and
  K.~Keutzer, ``Fbnet: Hardware-aware efficient convnet design via
  differentiable neural architecture search,'' in \emph{IEEE/CVF Conference on
  Computer Vision and Pattern Recognition (CVPR)}, 2019, pp. 10\,726--10\,734.

\bibitem{relu}
R.~H.~R. Hahnloser, R.~Sarpeshkar, M.~Mahowald, R.~J. Douglas, and H.~S. Seung,
  ``Digital selection and analogue amplification coexist in a cortex-inspired
  silicon circuit,'' \emph{Nature}, vol. 405, pp. 947--951, 2000.

\bibitem{elu}
D.-A. Clevert, T.~Unterthiner, and S.~Hochreiter, ``Fast and accurate deep
  network learning by exponential linear units (elus),'' \emph{arXiv preprint
  arXiv:1511.07289}, 2015.

\bibitem{MnasNet}
M.~Tan, B.~Chen, R.~Pang, V.~Vasudevan, M.~Sandler, A.~Howard, and M.~Le~QV,
  ``Platformaware neural architecture search for mobile. 2019 ieee,'' in
  \emph{IEEE/CVF Conference on Computer Vision and Pattern Recognition (CVPR)},
  2019, pp. 2815--2823.

\bibitem{RLNas}
B.~Zoph and Q.~V. Le, ``Neural architecture search with reinforcement
  learning,'' \emph{arXiv preprint arXiv:1611.01578}, 2016.

\bibitem{NAS-FPN}
G.~Ghiasi, T.-Y. Lin, and Q.~V. Le, ``Nas-fpn: Learning scalable feature
  pyramid architecture for object detection,'' in \emph{IEEE/CVF Conference on
  Computer Vision and Pattern Recognition (CVPR)}, 2019, pp. 7029--7038.

\bibitem{shapenet}
A.~X. Chang, T.~Funkhouser, L.~Guibas, P.~Hanrahan, Q.~Huang, Z.~Li,
  S.~Savarese, M.~Savva, S.~Song, H.~Su \emph{et~al.}, ``Shapenet: An
  information-rich 3d model repository,'' \emph{arXiv preprint
  arXiv:1512.03012}, 2015.

\bibitem{pytorch3D}
N.~Ravi, J.~Reizenstein, D.~Novotny, T.~Gordon, W.-Y. Lo, J.~Johnson, and
  G.~Gkioxari, ``Accelerating 3d deep learning with pytorch3d,'' \emph{arXiv
  preprint arXiv:2007.08501}, 2020.

\bibitem{marching}
W.~E. Lorensen and H.~E. Cline, ``Marching cubes: A high resolution 3d surface
  construction algorithm,'' in \emph{Proceedings of the 14th annual conference
  on Computer graphics and interactive techniques (SIGGRAPH)}, 1987, pp.
  163--169.

\end{thebibliography}

\end{document}